\documentclass[fleqn,12pt,a4paper]{article}
\usepackage{graphicx}
\usepackage{cite}

\newcommand{\bv}{\mbox{\bf v}}

\newcommand{\bA}{\mbox{\bf A}}
\newcommand{\bB}{\mbox{\bf B}}
\newcommand{\bC}{\mbox{\bf C}}
\newcommand{\bE}{\mbox{\bf E}}
\newcommand{\bG}{\mbox{\bf G}}

\newcommand{\bS}{\mbox{\bf S}}

\newcommand{\bV}{\mbox{\bf V}}

\newcommand{\graph}{\mbox{\bG=(\bV,\bE)}}

\newcommand{\proglang}[1]{{\fontseries{b}\selectfont #1}}
\newcommand{\pkg}[1]{{\fontseries{b}\selectfont #1}}

\long\def\symbolfootnote[#1]#2{\begingroup%
\def\thefootnote{\fnsymbol{footnote}}\footnote[#1]{#2}\endgroup}

\begin{document}

\begin{center}
\thispagestyle{empty}
\linespread{1.3}
{\LARGE High-dimensional Graphical Model Search with\\ \pkg{gRapHD} \proglang{R} Package}\\
\vspace{0.9cm}
\begin{large}
Gabriel C. G. de Abreu\symbolfootnote[1]{Department of Genetics and Biotechnology, Faculty of Agricultural Sciences, Aarhus University, DK-8830 Tjele, Denmark.}$^,$\symbolfootnote[2]{E-mail: Gabriel.Abreu@agrsci.dk}\hspace{0.8cm}
David Edwards$^{*,}$\symbolfootnote[3]{E-mail: David.Edwards@agrsci.dk}\hspace{0.8cm}
Rodrigo Labouriau$^{*,}$\symbolfootnote[4]{E-mail: Rodrigo.Labouriau@agrsci.dk}\\
\vspace{1cm}
\end{large}
\end{center}

\begin{abstract}
This paper presents the \proglang{R} package \pkg{gRapHD} for efficient selection of high-dimensional undirected graphical models. The package provides tools for selecting trees, forests, and decomposable models minimizing information criteria such as AIC or BIC, and for displaying the independence graphs of the models. It has also some useful tools for analysing graphical structures. It supports the use of discrete, continuous, or both types of variables.
\end{abstract}

\pagestyle{myheadings}

\markright{The \pkg{gRapHD} \proglang{R} package}

\section{Introduction}
We describe here the \proglang{R} package \pkg{gRapHD} which is designed to work with graphical models involving a large number of variables. These may be useful for modelling high dimensional networks in an wide range of biological applications ({\it e.g.} ecology \cite{dunne2002}; gene expression \cite{faith2007}; proteomics \cite{yosef2009}). Other applications are in computer sciences ({\it e.g.} internet \cite{dorogovtsev2003}), engineering ({\it e.g.} complex electronic circuits), physics \cite{dhamodaran2008}, sociology \cite{krause2007}.

The graphical models used here are classes of multivariate distributions whose conditional independence properties are encoded by a graph in the following way. The random variables are represented as vertices (nodes), and two vertices are connected by an edge (line) when the corresponding variables are not conditionally independent given the other variables represented in the graph. Thus the absence of an edge connecting two vertices indicates conditional independence of the two corresponding variables given the other variables.

This type of model has been used in high dimensional contingency tables ({\it e.g.} log-linear mo\-dels \cite{goodman1973,darroch1980}), continuous variables ({\it e.g.} the covariance selection model described by \cite{dempster1972}), and models containing both continuous and discrete variables \cite{lauritzen1989}. Modern accounts of graphical models can be found in \cite{edwards2000}, \cite{lauritzen1996}, and \cite{whittaker1990}.

The use of graphical models for large numbers of variables can be very challenging, both because of computational limitations, and because of intrinsic statistical difficulties (particularly when the sample sizes are small). Consequently the use of such models has often been restricted to small problems. The package \pkg{gRapHD} presented here is intended for high-dimensional graphical modelling. The central functions, \verb@minForest@ and \verb@stepw@, search respectively for the optimal forest and the optimal decomposable model, where
optimality is typically defined in termed of an information criterion (AIC or BIC).

The the \pkg{gRapHD} package is here presented using three distinct examples, which are described in Section \ref{sec:examples}. The basic definitions and notations used throughout the paper are found in Section \ref{sec:basic}, and the structure of the \verb@gRapHD@ object class defined by the package is described in Section \ref{sec:object}. Sections \ref{sec:search} to \ref{sec:plot} present the functions in the package, using the examples previously introduced.

\section{Three examples}
\label{sec:examples}
The features of the \pkg{gRapHD} package will be presented using three different examples covering continuous, discrete, or both types of variables (examples 1, 2, and 3, respectively). We here describe the data and show the graphical model selected.

\subsection{Example 1 - continuous: Periodontitis data}
The data arise from a functional genomics study in gingival tissue \cite{demmer2008}. The study investigates the differences in the gene expression profiles of interproximal papillae tissues of diseased (periodontitis) and health sites. Ninety healthy non-smoker patients with moderate to advanced periodontitis were used in this study. Each patient contributed with at least two diseased samples and a healthy papilla, if available. A total of 247 samples were collected (64 from healthy sites and 183 from diseased sites). The transcription profiles of the samples were evaluated using Affymetrix human genome arrays with 54,675 probe sets.

Only 64 arrays from independent diseased sites were considered. The data were pre-processed using \verb@justRMA@ from the \pkg{affy} package \cite{gautier2004}. To reduce the amount of data, probes with variance $< 0.62$ were omitted from the analysis. Thus our dataset is composed of 1,545 probes, from 64 different patients. The objective is to characterize the gene co-expression network in patients with periodontitis. The model selected by the \verb@stepw@ function can be seen in Figure~\ref{fig:final} (A).

\subsection{Example 2 - discrete: HapMap data}
The goal of the International HapMap Project is to characterize human genetic variation \cite{HapMap2003}. The project recorded differences in the sequence of bases that composes the DNA, the {\it SNPs} ({\it S}ingle {\it N}ucleotide {\it P}olymorphisms). There are four different bases (A, T, G, and C) that can occur at each position in the DNA ({\it locus}), and if different individuals have different bases at a locus, the locus can be considered polymorphic (provided that none of the alleles are too rare). Each different base occurring in one locus is considered as an allele.

\begin{figure}[!ht]
\centering
\includegraphics[height=5.1cm,width=5.1cm]{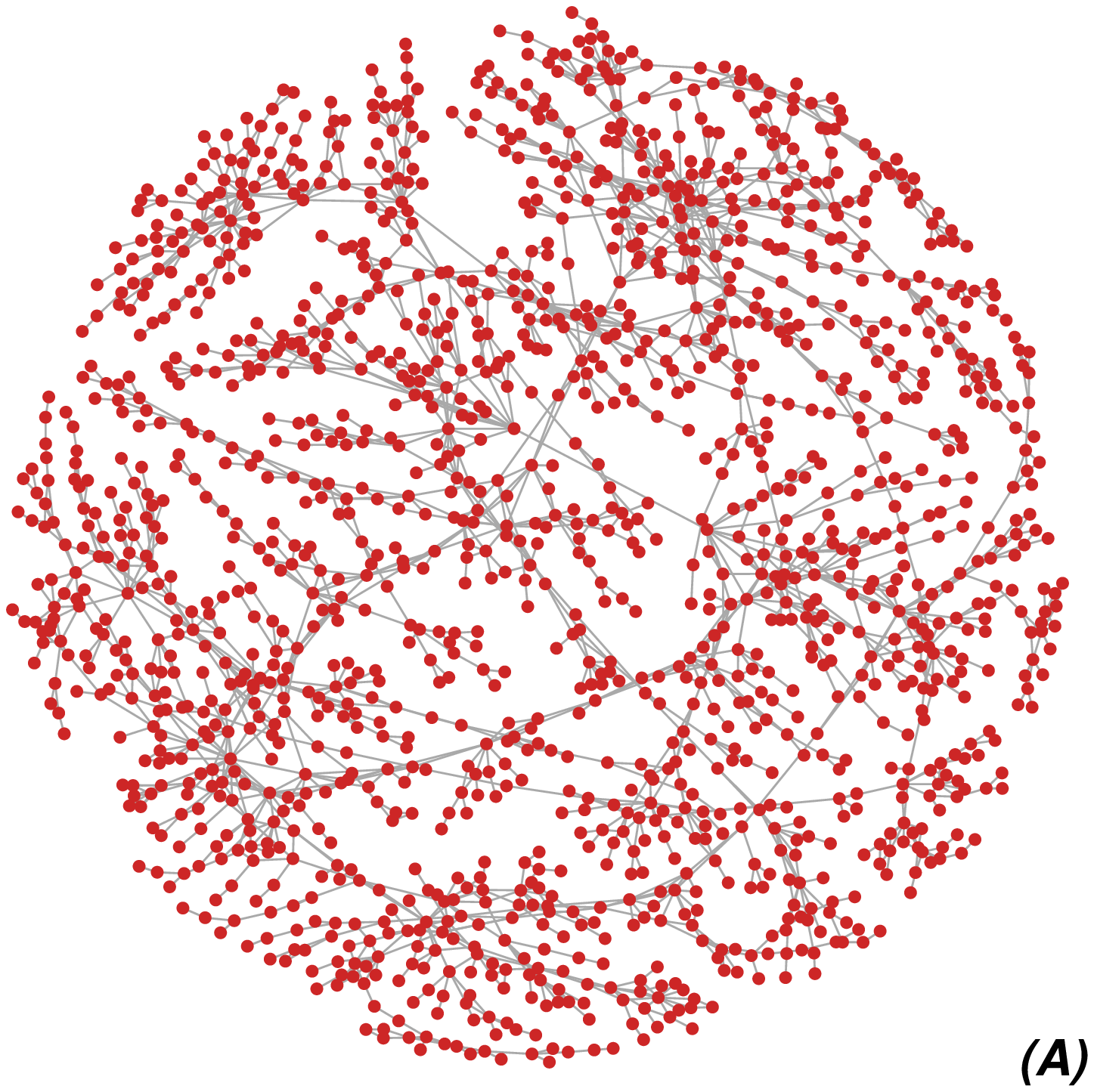}
\includegraphics[height=5.1cm,width=5.1cm]{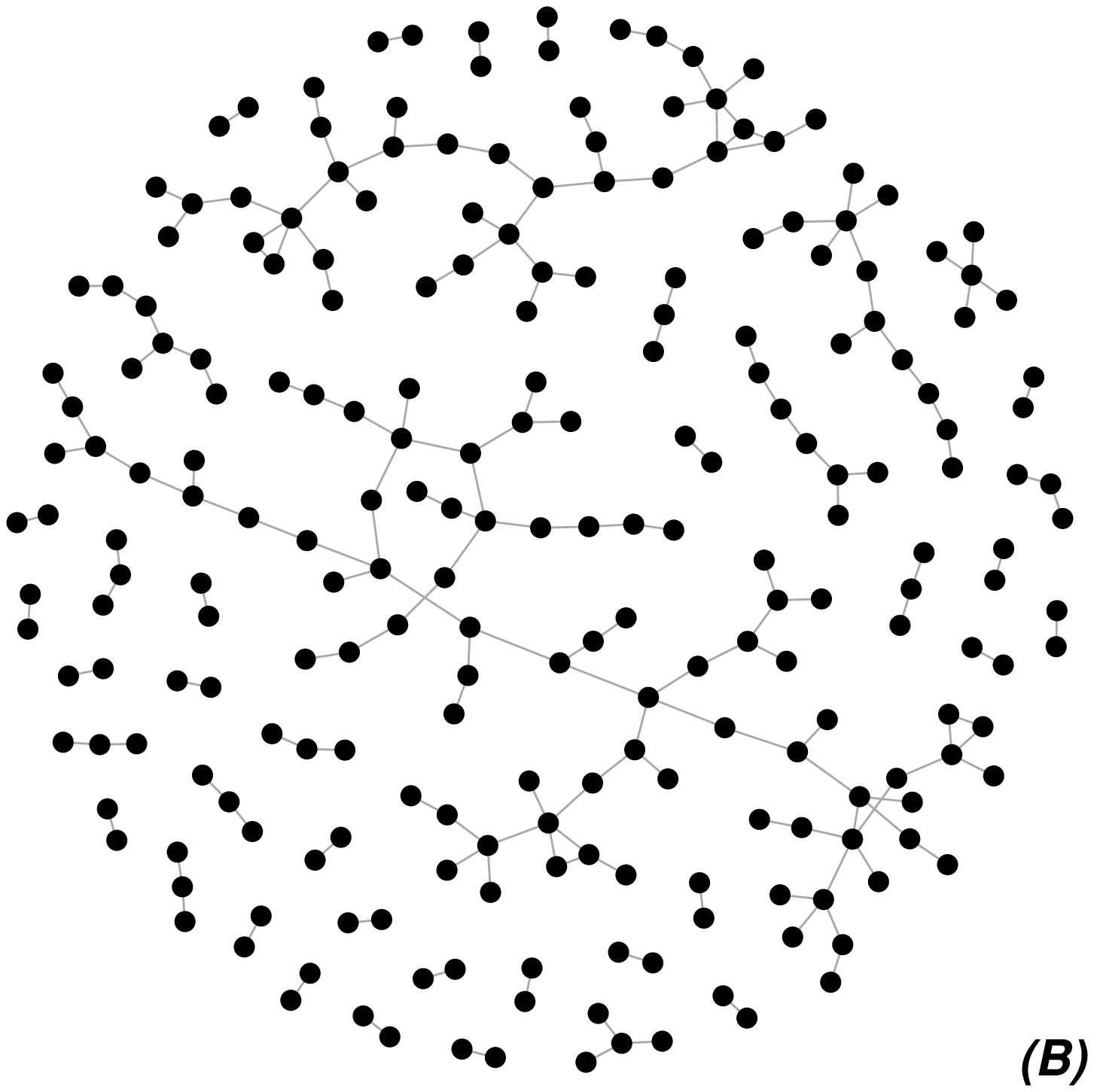}
\includegraphics[height=5.1cm,width=5.1cm]{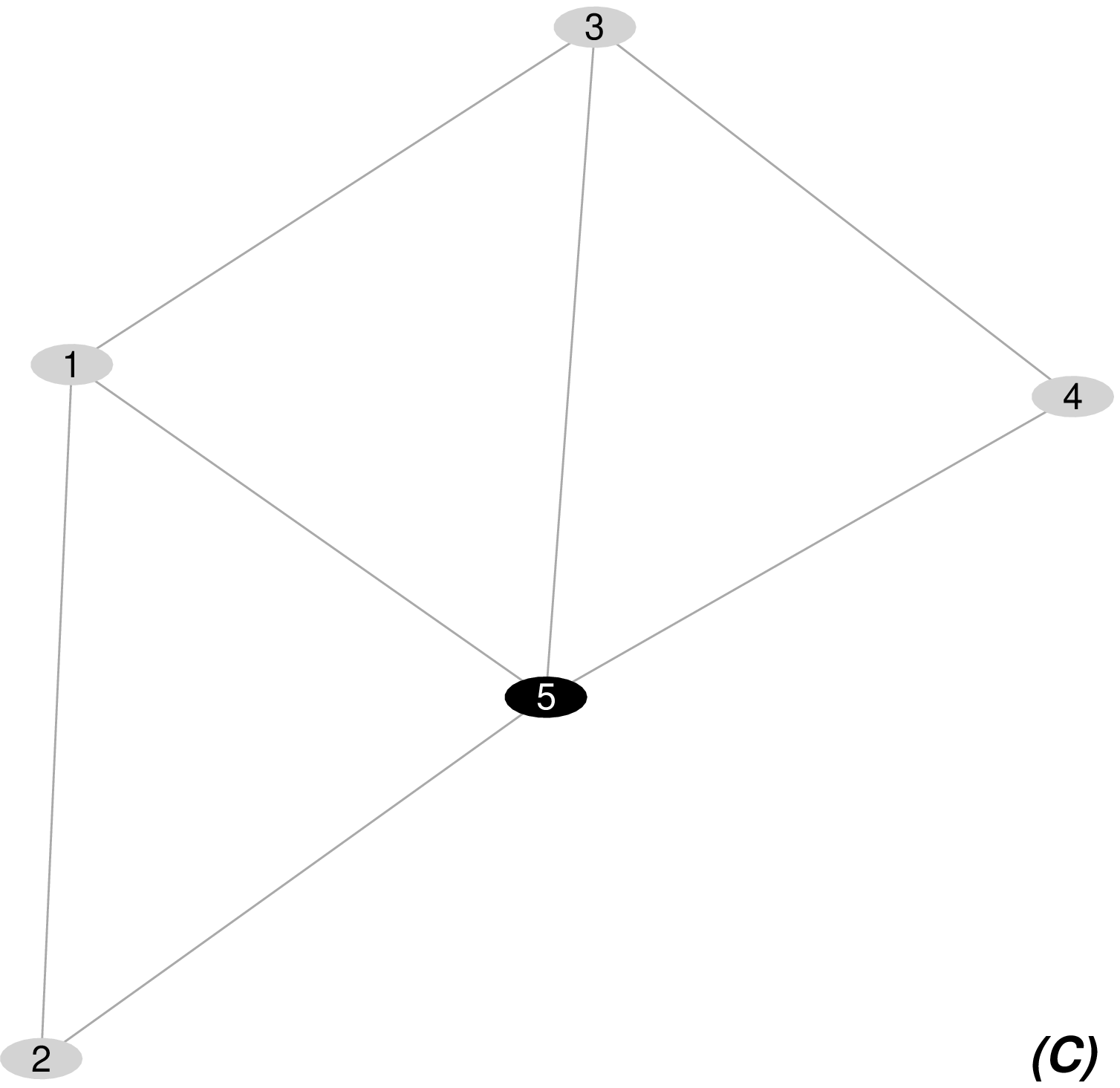}
\caption{{\bf Three inferred networks}. (A) Periodontitis data (example 1); (B) HapMap data (example 2); (C) Iris data (example 3).}
\label{fig:final}
\end{figure}

We use here only the polymorphic SNPs from the Yoruba population (Ibadan, Nigeria - West Africa) with complete data. From this population,
we selected the chromosome 17, representing 606 SNPs in 176 individuals, for synonymous coding SNPs (SNPs which even with different structure code the same amino acid) with minor allele frequency greater than 0.25 (download on the $3^{rd}$ June 2009 from the HapMart website - {\tt www.hapmap.org}). After eliminating 9 individuals with very high missing values percentage we obtained 334 loci without missing values. Using the information of the stated reference allele we codified the genotypes of the individuals as: ``0" for homozygous wild type (individual with the reference allele in both DNA strands); ``1" for heterozygous (individual with only one copy of the reference allele); and ``2" for homozygous mutant (individual with no copy of the reference allele). The objective is to determine the relationship between different loci through the network structure. The optimal decomposable representation of the network (minimum BIC) is displayed in Figure~\ref{fig:final} (B).

\subsection{Example 3 - mixed: Iris data}
The iris flower data were introduced by Anderson in 1935, but is also known as the Fisher's iris data (1936). A sample of 150 plants, 50 from each of three species ({\it Iris setosa}, {\it Iris versicolor}, and {\it Iris virginica}), had the sepal and petal lengths and widths recorded. The objective is to describe the structure of correlation between the different measurements considering the 3 different species. The final model can be seen in Figure~\ref{fig:final} (C).

\section{Basic definitions and notation}
\label{sec:basic}
\begin{figure}[!ht]
\centering
\includegraphics[width=.33\linewidth]{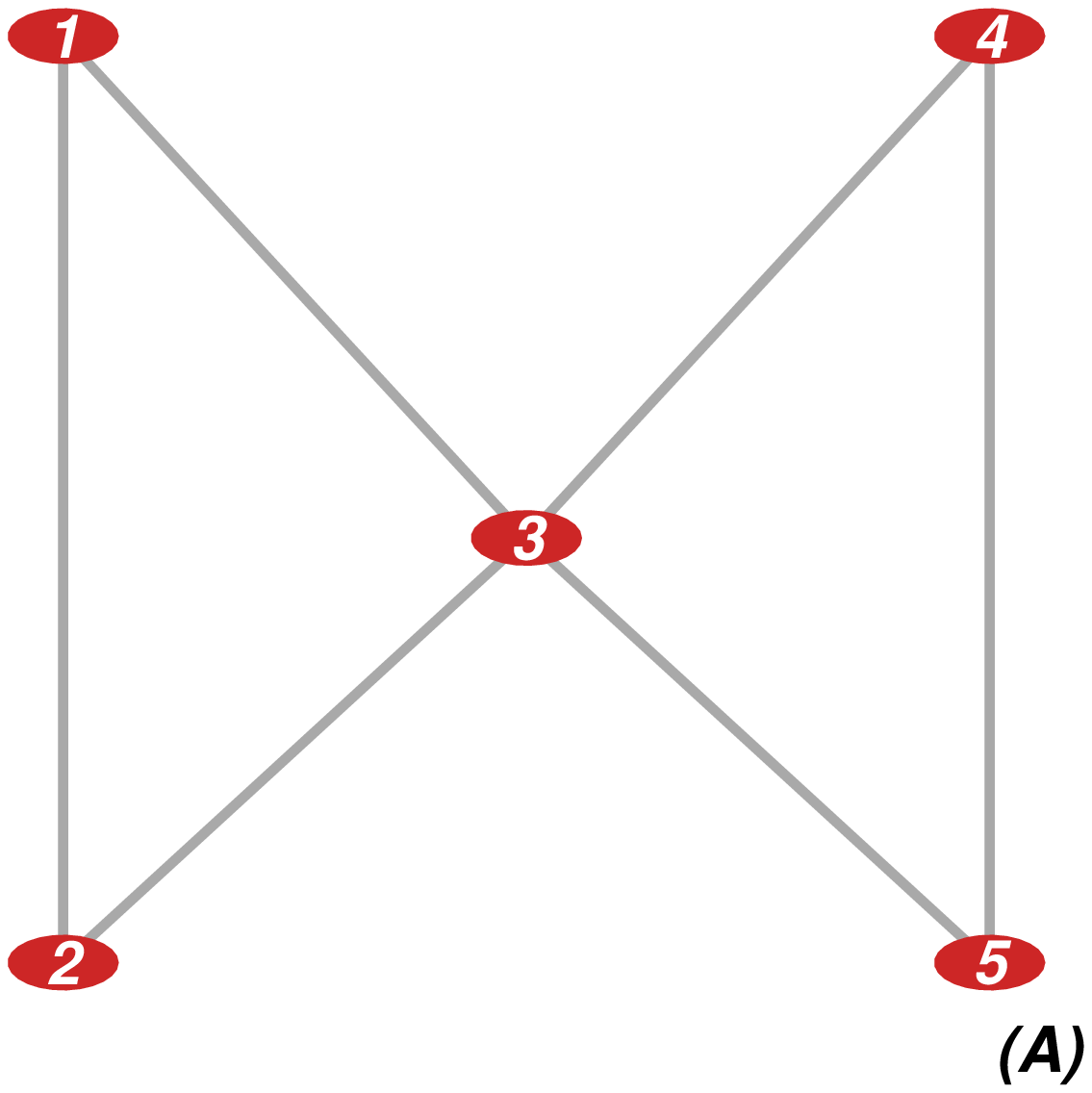} \hspace{2cm}
\includegraphics[width=.33\linewidth]{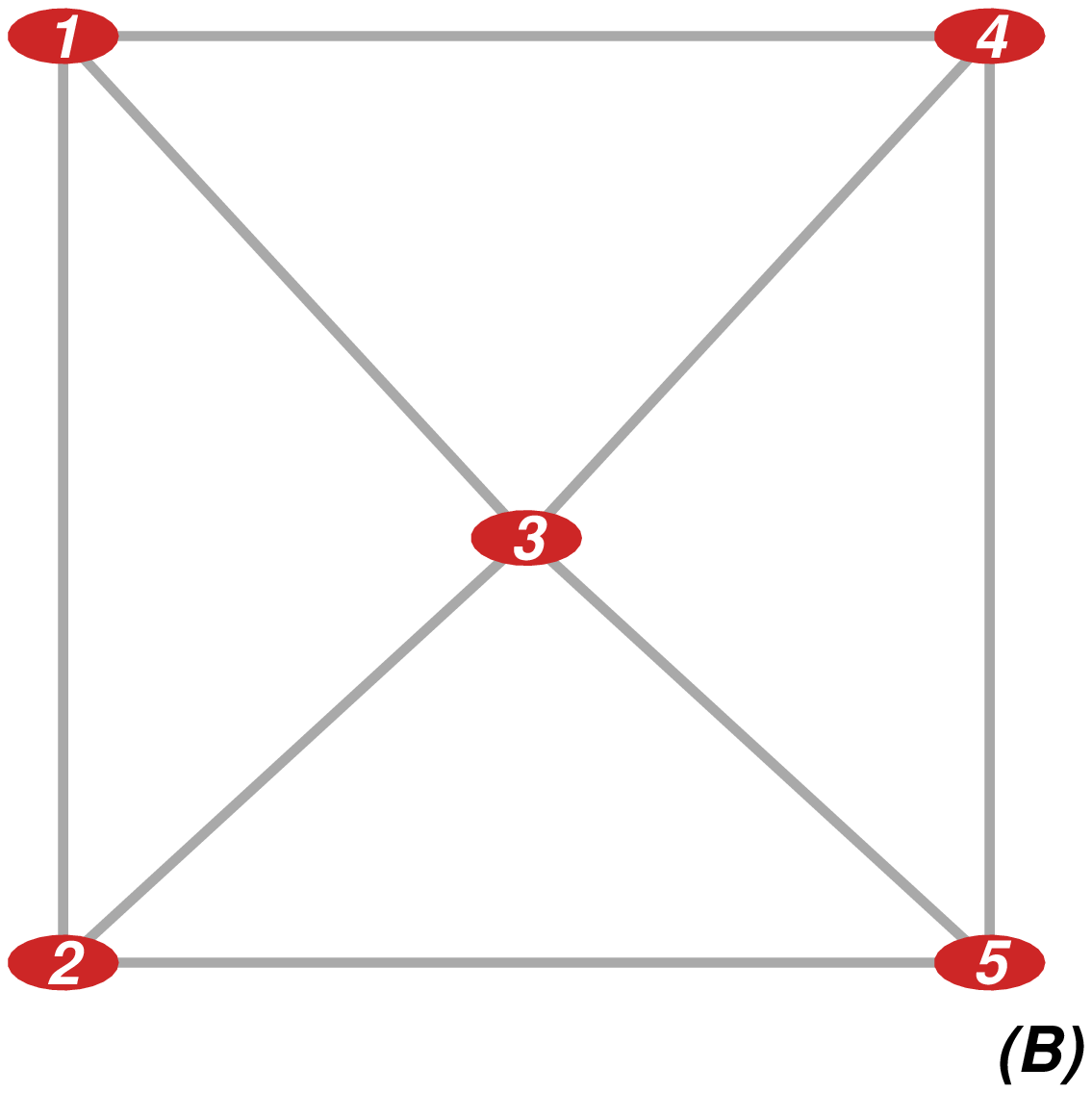}
\caption{{\bf Example of triangulation}. (A) The structural relationship between the 5 vertices indicates that vertices 1 and 2 are conditionally independent of vertices 4 and 5, given vertex 3. (B) Adding the edges (1,4) and (2,5) renders the graph non-triangulated.}
\label{fig:example}
\end{figure}

We here give a brief sketch of the theory of graphical models. For a more complete account see for example \cite{lauritzen1996}. Graphical models combine graph theory and probability theory. Each vertex represents a random variable, and two vertices are connected when
they are not conditionally independent given the remaining variables. For example, from the graph presented in Figure~\ref{fig:example}~(A), we see that the variables 1 and 2 are not conditionally independent given variables 3, 4 and 5, but that variables 1 and 4 are conditionally independent given the variables 2, 3 and 5.

Define a graphical model as an undirected graph $\graph$, where $\bV=\{v_1,...,v_p\}$ is the set of $p$ vertices ($p$ finite), and $\bE$ is the set of edges, a subset of $\bV\times\bV$ (unordered pairs), where multiple edges and self-loops are not allowed. Furthermore, an edge $e=(u,v)\in\bE$ indicates that the variables associated to $u$ and $v$ are not conditionally independent given all the other variables. If we are only interested in a subset of such relations, we can define a subgraph of $\bG$ as $\bG_A=(\bA,\bE_A)$, where $\bA\subseteq\bV$ and $\bE_A\subseteq\bE$ \cite{bollobas2000}. For example, in Figure~\ref{fig:example}~(A), $\bV=\{1,2,3,4,5\}$ and $\bE=\{(1,2),(1,3),(2,3),(3,4),(3,5),(4,5)\}$, and $\bA=\{1,2,3\}$ renders the subgraph to the graph $\bE_A=\{(1,2),(1,3),(2,3)\}$.

A graph is complete if every pair of vertices is connected by an edge. If a subgraph is maximally complete, it is called a {\it clique}: in this case the addition of any other vertex would renders the subgraph incomplete. In Figure~\ref{fig:example}~(A), $\{1,2,3\}$ and $\{3,4,5\}$ are cliques. 

In a graph $\bG$, two vertices, $u$ and $v$, are said to be connected if there is a sequence $u=v_1,...,v_k=v$ of distinct vertices such that $(v_{i-1},v_i)\in\bE,\ \forall\ i=2,...,k$. The sequence $u=v_1,...,v_k=v$ is called {\it path}. In the Figure~\ref{fig:example}~(A), the vertices 1 and 5 are connected since there exists a path between them, for example $1,2,3,5$. A {\it cycle} is a path which the end vertices are the same ($u=v$), as $1,2,3,1$ in Figure~\ref{fig:example}~(A). A cycle $u=v_1,...,v_k=u$ is {\it chordless} if $v_i$ and $v_j$ are only connected by an edge when $|i-j|=1$. A graph is called {\it triangulated} if it has no chordless cycles of length greater than three. For example, the graph in Figure~\ref{fig:example}~(A) is triangulated, but the graph in Figure~\ref{fig:example}~(B) is not, since the cycle $1,2,5,4,1$ of length four is chordless.

A subset $\bC\subseteq\bV$ separates two disjoint subsets of $\bV$, $\bA$, and $\bB$, if all paths from $v\in\bA$ to $u\in\bB$ pass through $\bC$. In addition, a triple $(\bA,\bB,\bC)$ of disjoint subsets of $\bV$ decomposes the graph $\graph$ if: (1) $\bV=\bA\cup\bB\cup\bC$, (2) $\bC$ separates $\bA$ from $\bB$, and (3) $\bC$ is complete. This definition implies that a graph is {\it decomposable} if it is complete, or exists a decomposition $(\bA,\bB,\bC)$, with $\bA\neq\emptyset$ and $\bB\neq\emptyset$, into decomposable subgraphs $\bG_{A\cup C}$ and $\bG_{B\cup C}$. A graph $\graph$ is decomposable if and only if it is triangulated \cite{lauritzen1996}. The example shown in Figure~\ref{fig:example}~(A) is decomposable, with $A=\{1,2\}$, $B=\{4,5\}$, and $C=\{3\}$.

The cliques $C_1,...,C_k$ in a triangulated graph can be numbered in such a way that for all $j=1,...,k$, $H_j=C_i\cup...\cup C_j$, $R_j=C_j\backslash H_{j-1}$, and $S_j=H_{j-1}\cap C_j$ gives that (1) for all $i>1$ there is a $j<i$ such that $S_i\subseteq C_j$, and (2) the sets $S_i$ are complete for all $i$ \cite{lauritzen1996}. This sequence of cliques is called a {\it perfect sequence}. The sets $H_j$, $R_j$, and $S_j$ are named histories, residuals, and separators, respectively. A perfect sequence of the graph in Figure~\ref{fig:example}~(A) is $\{2,3,1,4,5\}$, with cliques $C_1=\{1,2,3\}$ and $C_2=\{3,4,5\}$, respective separators $S_1=\emptyset$ and $S_2=\{3\}$, histories $H_1=\{1,2,3\}$ and $H_2=\{1,2,3,4,5\}$, and residuals $R_1=\{1,2,3\}$ and $R_2=\{4,5\}$.

A key property of decomposable graphs is that the probability densities of such models can be factorized as:
$$f(\bv)=\frac{\prod_{C\in\cal{C}}f(\bv_{C})}{\prod_{S\in\cal{S}}f(\bv_{S})^{\nu(S)}},$$ 
where $\cal{C}$ is the class of cliques in a perfect sequence, and $\nu(\bS)$ is the number of times that $\bS$ occurs as a separator in this perfect sequence (possibly including the empty set) \cite{lauritzen1996}. Furthermore, these models have an explicit formula for the maximum likelihood estimators.

\begin{figure}[!ht]
\centering
\includegraphics[width=.33\linewidth]{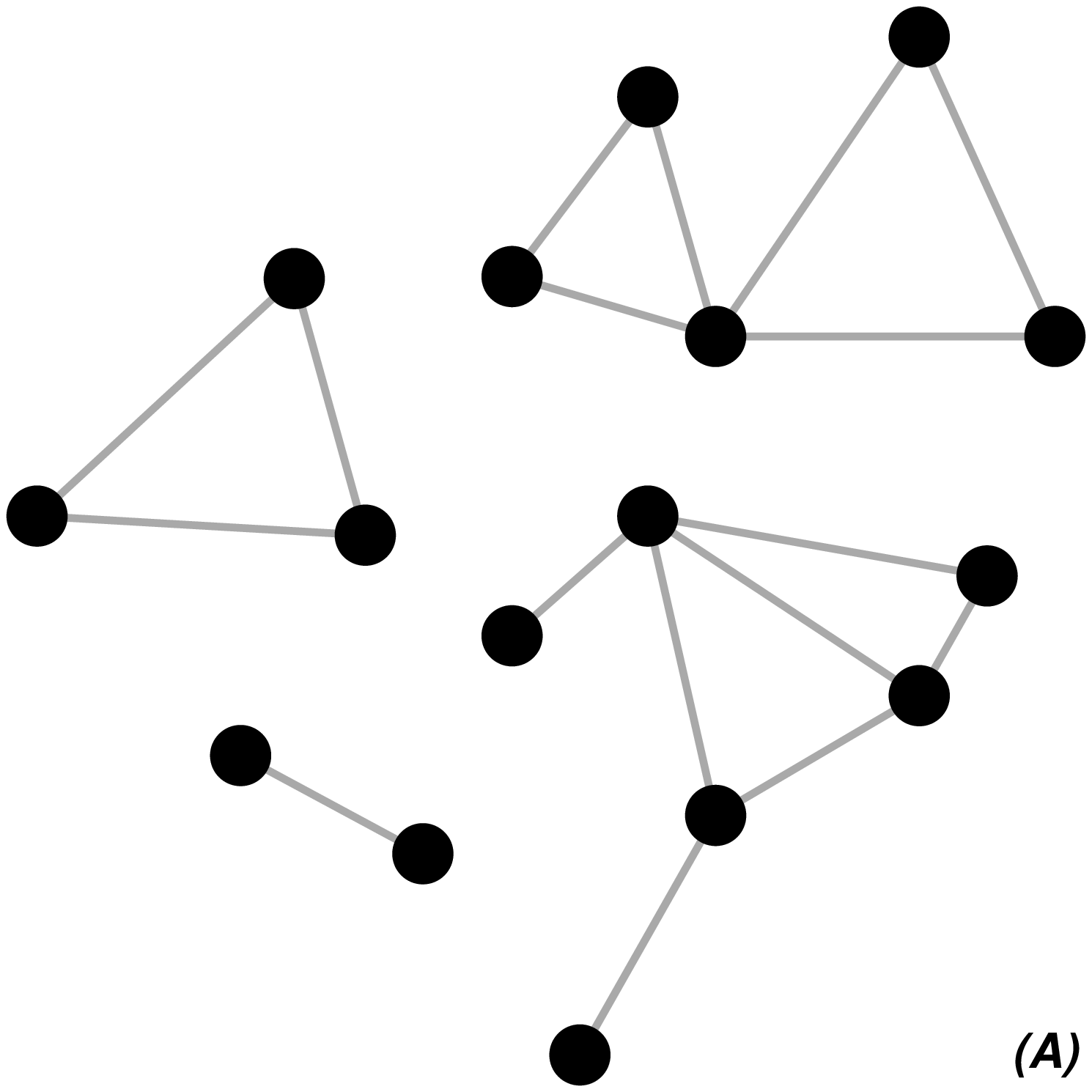} \hspace{2cm}
\includegraphics[width=.33\linewidth]{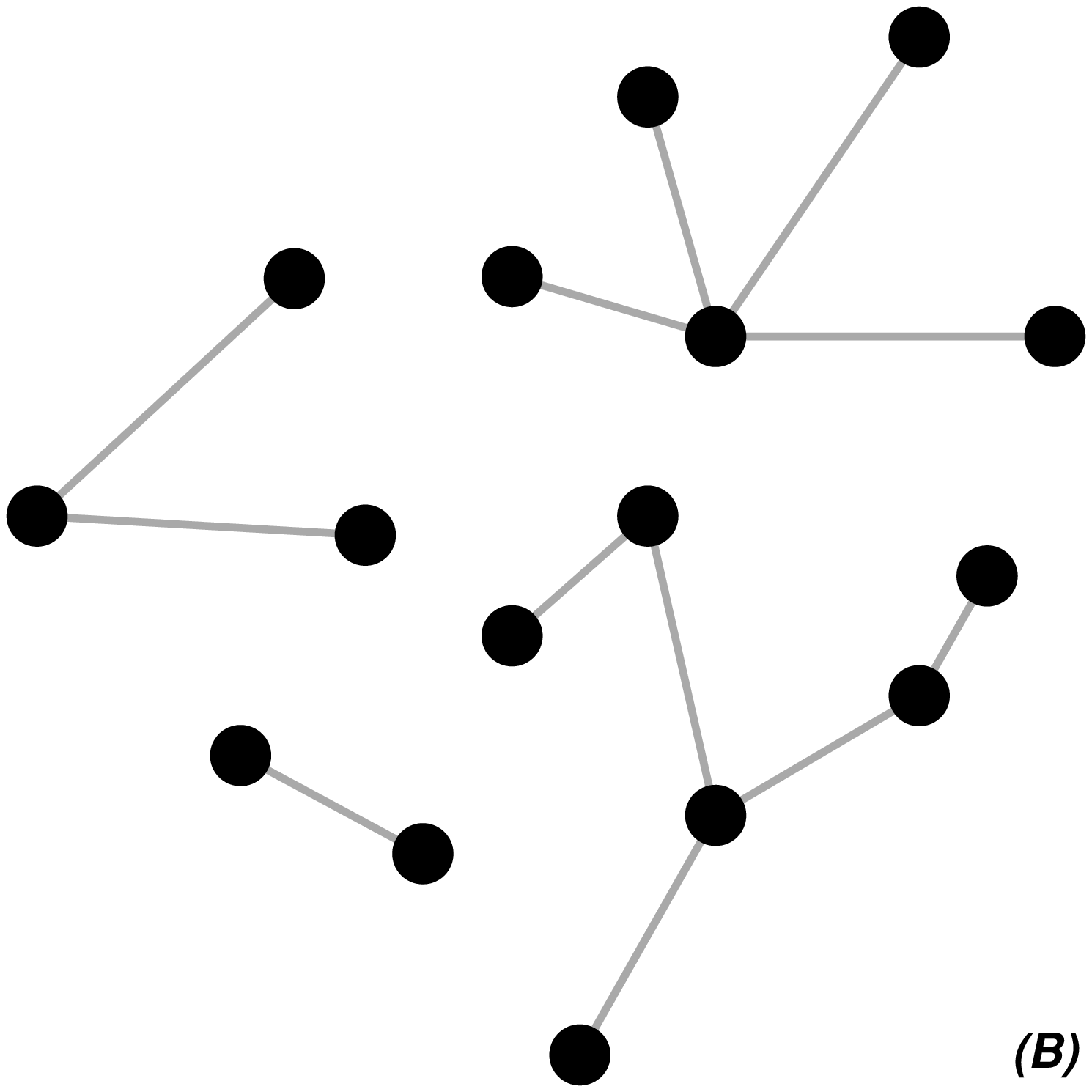}
\caption{{\bf Example tree and forest}. (A) Graph with four connected components. (B) One possible spanning forest (each connected component is a tree) for the graph in (A).}
\label{fig:forest}
\end{figure}

A {\it forest} is a graph containing no cycles. It may be composed of several connected components called trees, i.e. a tree is a connected acyclic graph \cite{bondy2008}. Given a set of edge weights, a {\it minimum spanning forest} $\bG_m=(\bV_m,\bE_m)$ of a graph $\graph$ is a forest with $\bV_m=\bV$ and $\bE_m\subset\bE$ that has minimum sum of edge weights, among all possible such forests. Examples of edge weights are the contribution of each edge to the BIC or the contribution to minus the log-likelihood. Figure~\ref{fig:forest}~(A) shows a graph with four connected components, and Figure~\ref{fig:forest}~(B) shows a spanning forest of this graph.

\begin{figure}[!ht]
\centering
\includegraphics[width=.33\linewidth]{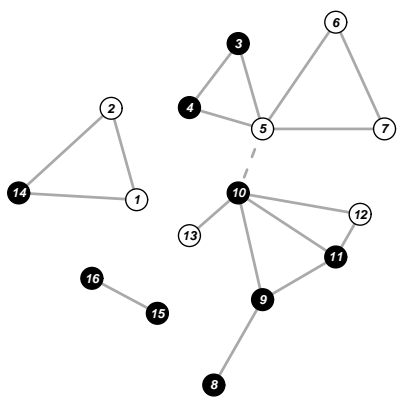} \hspace{2cm}
\caption{{\bf Forbidden path}. Continuous variables are represented as circles, and discrete variables as dots. The edge (5,10) creates a forbidden path in the graph, since a continuous variable (variable 5) connects non-adjacent discrete variables (variables 3 and 4 to variables 8, 9, 10, and 11).}
\label{fig:forbidden}
\end{figure}

The variables in the model can be discrete, continuous, or both (mixed). In the first case, in which each variable assumes a value in a set of levels, the models are based on the multinomial distribution. In the second case, the models are based on the multivariate Gaussian distribution. In the mixed case the CG (conditional Gaussian) distribution is assumed; the variances can be homogeneous or heterogeneous across different levels of the discrete variables. Furthermore, a mixed model is strongly decomposable when its graph is triangulated and no forbidden paths occur. A {\it forbidden path} is a path between two non-adjacent discrete vertices passing through only continuous vertices, as showed in Figure~\ref{fig:forbidden}. For more details see \cite{lauritzen1996} pages 7-12.

In this paper we describe the main features of the new \proglang{R} package \pkg{gRapHD}, and how it can be used for graphical model search. Most of the core functions are programmed in \proglang{ANSI C}, due to its computational efficiency. The package as well as the code are available from {\tt http://CRAN.R-project.org/}.

\section[The gRapHD object]{The {\tt gRapHD} object}
\label{sec:object}
The \pkg{gRapHD} package groups a number of functions designed for efficient selection of high-dimensional undirected graphical models. The set of variables can contain only discrete, continuous, or both types. All the information about the model is stored in a \verb@gRapHD@ object which contains the following elements:
\begin{itemize}
    \item \verb@edges@: Matrix with 2 columns, each row representing one edge, and each column one of the vertices in the edge.
    \item \verb@p@: Number of variables (vertices) in the model.
    \item \verb@stat.minForest@: Measure used (LR, AIC, BIC, or a user defined function) by the \verb@minForest@ function.
    \item \verb@stat.stepw@: Measure used (LR, AIC, BIC, or a user defined function) by the \verb@stepw@ function.
    \item \verb@statSeq@: Numeric vector with the value of the \verb@stat.minForest@ for each edge found by \verb@minForest@, or the change in the \verb@stat.stepw@ for each edge found by \verb@stepw@.
		\item \verb@vertNames@: Vector with the original vertices' names. If no names are attributed, then the vertices will be named according to their original order in the data.
		\item \verb@numCat@: Vector with number of levels for each variable (0 if continuous).
    \item \verb@homog@: \verb@TRUE@ if the covariance is homogeneous (only used in the mixed case).
    \item \verb@numP@: Vector with the number of estimated parameters for each edge.
    \item \verb@minForest@: Vector of length 2, with the row indexes of the first and last edges found by the \verb@minForest@ function.
    \item \verb@stepw@: Vector of length 2, with the row indexes of the first and last edges found by the \verb@stepw@ function.
\end{itemize}

The \verb@gRapHD@ object, besides being the result of a model search, functions as an input parameter in most other functions in the package. A null model can be created using the function \verb@as.gRapHD@, as in
\begin{verbatim}
R> m <- as.gRapHD(NULL)
List of 9
 $ edges    : int[0 , 1:2]
 $ p        : int 0
 $ stat.user: chr "LR"
 $ statSeq  : num(0)
 $ numCat   : int(0)
 $ homog    : logi TRUE
 $ numP     : num(0)
 $ vertNames: logi NA
 $ userDef  : num [1:2] 0 0
 - attr(*, "class")= chr "gRapHD"
\end{verbatim}

\section{Search for graphical models}
\label{sec:search}
The core functions in the package are \verb@minForest@ and \verb@stepw@. The \verb@minForest@ function finds a minimal AIC or BIC forest, or the maximum likelihood tree for the data. The \verb@stepw@ function performs forward search for the triangulated graph that minimises a given measure. The minimized measure used can be either {\tt -LR} (likelihood-ratio), {\tt BIC} (Bayesian Information Criteria), {\tt AIC} (Akaike Information Criteria), or a function specified by the user.

All calculations use the data set specified by the \verb@dataset@ parameter, which holds the raw data, in \verb@dataframe@ format, with the column storage mode defining its type (discrete or continuous). The discrete variables are defined as factors, and the continuous as numerical. In the first case, there should be more than one level (otherwise there is no gain in using such variable), and all levels must be represented in the sample. If all variables are continuous, the dataset can be given as a numeric matrix. Missing values (\verb@NA@) are not allowed. The variables are referred by their indexes in the \verb@vertNames@ attribute. In this way, the edge representation in the \verb@gRapHD@ object is in the format $(v_1,v_2)$, where $v_1$ and $v_2$ are the indexes in \verb@vertNames@, and $v_1<v_2$. For example, the Iris data is presented as
\begin{verbatim}
R> str(iris)
'data.frame':   150 obs. of  5 variables:
 $ Sepal.Length: num  5.1 4.9 4.7 4.6 5 5.4 4.6 5 4.4 4.9 ...
 $ Sepal.Width : num  3.5 3 3.2 3.1 3.6 3.9 3.4 3.4 2.9 3.1 ...
 $ Petal.Length: num  1.4 1.4 1.3 1.5 1.4 1.7 1.4 1.5 1.4 1.5 ...
 $ Petal.Width : num  0.2 0.2 0.2 0.2 0.2 0.4 0.3 0.2 0.2 0.1 ...
 $ Species     : Factor w/ 3 levels "setosa","versicolor",..: 1 1 1 ...
\end{verbatim}
The discrete variable (\verb@Species@) is identified as a factor with three levels. All other variables are continuous (as indicated by the numerical type `\verb@num@'). The columns names (\verb@Sepal.Length@, \verb@Sepal.Width@, \verb@Petal.Length@, \verb@Petal.Width@, and \verb@Species@) are stored in the \verb@vertNames@ attribute of the \verb@gRapHD@ object. The variables are always referred to as the original column number in the \verb@dataset@, {\it e.g.} the variable \verb@Sepal.Width@ is variable 2, while \verb@Species@ is variable 5. The edge connecting \verb@Sepal.Width@ and \verb@Species@ is represented by a row
in the \verb@edges@ attribute consisting of the vector \verb@c(2,5)@.

\subsection{Search for minimum spanning forests}
The function \verb@minForest@ searches for an optimal tree or forest using the algorithm of \cite{chow1968}. If the ML measure is used, the function returns a tree, but if the AIC or BIC is used, the function may return a forest or a tree. Per default the BIC measure is used. Starting from an empty edge set, the algorithm repeatedly adds the edge that optimizes the selected measure. Only edges that preserve the tree/forest structure can be added, i.e., no cycles can be generated. The procedure continues until no more edges can be added. For example, if the selected measure is the BIC, at first the pairwise values are computed and in each step the edge that reduces the most the total BIC is added, if it exists. The algorithm is
	\begin{description}
		\item[Step 1:] Calculate the BIC for all possible edges.
		\item[Step 2:] Select the edge that improves the most the model's BIC.
		\item[Step 3:] If there is no such edge, stop.
		\item[Step 4:] Test if the addition of this edge creates a cycle or a forbidden path.
		\item[Step 5:] If it does, select the next edge with best improvement and return to {\bf Step 3}.
		\item[Step 6:] Add the edge to $\bE$, remove it from the list of possible edges, and return to {\bf Step 2}.
	\end{description}
	
For mixed models, the algorithm finds the strongly decomposable forest that minimizes the selected measure \cite{edwards2010}.

For the periodontitis data, the summary of the minimum spanning forest is presented below. The graph found is actually a tree and not a forest, since all vertices are connected ($p-1$ edges). All variables are continuous, and the measure minimised was the BIC (default).

\begin{verbatim}
R> library("gRapHD")
R> periodontitisForest <- minForest(periodontitisData)
gRapHD object
Number of edges       = 1544
Number of vertices    = 1545
Model                 = continuous
Statistic (minForest) = BIC
Edges from minForest  = 1...1544
\end{verbatim}

\subsection{Forward search}
The function \verb@stepw@ searches for decomposable models minimising a given measure by adding edges to a initial model, also decomposable. The algorithm is also iterative, determining at each step the add-eligible edges, i.e., the edges that if added preserve the triangularity. Among these edges, the one that reduces the most the selected measure is added to the graph. The algorithm, showed below, stops when no more add-eligible edges are found. The structure of components in the starting model is preserved as default, which means that if the function starts from a forest with $k$ isolated components, the final model will also have $k$ isolated components. This can be changed setting the option \verb@join@ to \verb@TRUE@.
	\begin{description}
		\item[Step 1:] Calculate the BIC for all add-eligible edges.
		\item[Step 2:] Select the edge that improves the most the model's BIC.
		\item[Step 3:] If there is no such edge, stop.
		\item[Step 4:] Add the edge to $\bE$, and return to {\bf Step 1}.
	\end{description}

For example, if the function is applied on the tree found in the previous section, 999 iterations are necessary (998 edges are added to the tree).

\begin{verbatim}
R> periodontitisForward <- stepw(periodontitisForest,periodontitisData)
gRapHD object
Number of edges       = 2542
Number of vertices    = 1545
Model                 = continuous
Statistic (minForest) = BIC
Statistic (stepw)     = BIC
Edges from minForest  = 1...1544
Edges from stepw      = 1545...2542
\end{verbatim}

The default measure is the BIC. We can see that in the final model, the first 1,544 edges were found by \verb@minForest@ (edges 1 to 1,544), and the last 998 by \verb@stepw@ (edges 1,545 to 2,542). The final graph can be seem in Figure~\ref{fig:final} (A).

The algorithm can start from an empty model (which is always decomposable). If we use the iris data, the final model (Figure~\ref{fig:final} C) found is the same as if it had started from the forest/tree graph. The code for it is showed below. Initially a \verb@gRapHD@ object is generated, from an empty model with 5 variables, being the first 4 continuous and the last discrete with 3 levels. The model is considered heterogeneous. As we are starting from 5 isolated components, they are allowed to be joined.

\begin{verbatim}
R> irisEmpty <- as.gRapHD(matrix(integer(0),,2),p=5,numCat=c(0,0,0,0,3),
+                         homog=FALSE)
R> irisForward <- stepw(irisEmpty,iris,join=TRUE)
\end{verbatim}

\subsection{Computational performance}
The performance of \pkg{gRapHD} was evaluated using a Intel(R) Xeon(R) CPU E5450 3.00GHz with 31Gb of RAM, running Linux 64 bits. The memory use and CPU time of the three examples are presented in Table~\ref{tab:perf}.
\begin{table*}[!ht]
	\centering
  \begin{footnotesize}
		\begin{tabular}{l|c|c|c|c|c|c|c}
			\hline
			\hline
			Example & Number of & \multicolumn{2}{c|}{Number of edges added} & \multicolumn{2}{c|}{CPU time (sec.)}&\multicolumn{2}{c}{Memory (Mb)}\\
			\cline{2-8}
			& vertices & \verb@minForest@ & \verb@stepw@ & \verb@minForest@ & \verb@stepw@ & \verb@minForest@ & \verb@stepw@ \\
			\hline
			Periodontitis & 1,545 & 1,544 & 998 & 9.27 & 2024.14 & 36.64 & 0.0956\\
			\hline
			HapMap & 334 & 189 & 5 & 0.18 & 2.18 & 1.66 & 0.01 \\
			\hline
			Iris & 5 & 4 & 3 & 0.001 & 0.003 & 0.002 & 0.0003\\
			\hline
			\hline
		\end{tabular}
	\caption{{\bf Computational performance}. CPU time (for the whole procedure) and memory use of the three examples.}
	\label{tab:perf}
  \end{footnotesize}
\end{table*}

We see that the \verb@minForest@ function is highly efficient in terms of CPU time but may require substantial memory.

\begin{figure}[!ht]
\centering
\includegraphics[width=.48\linewidth]{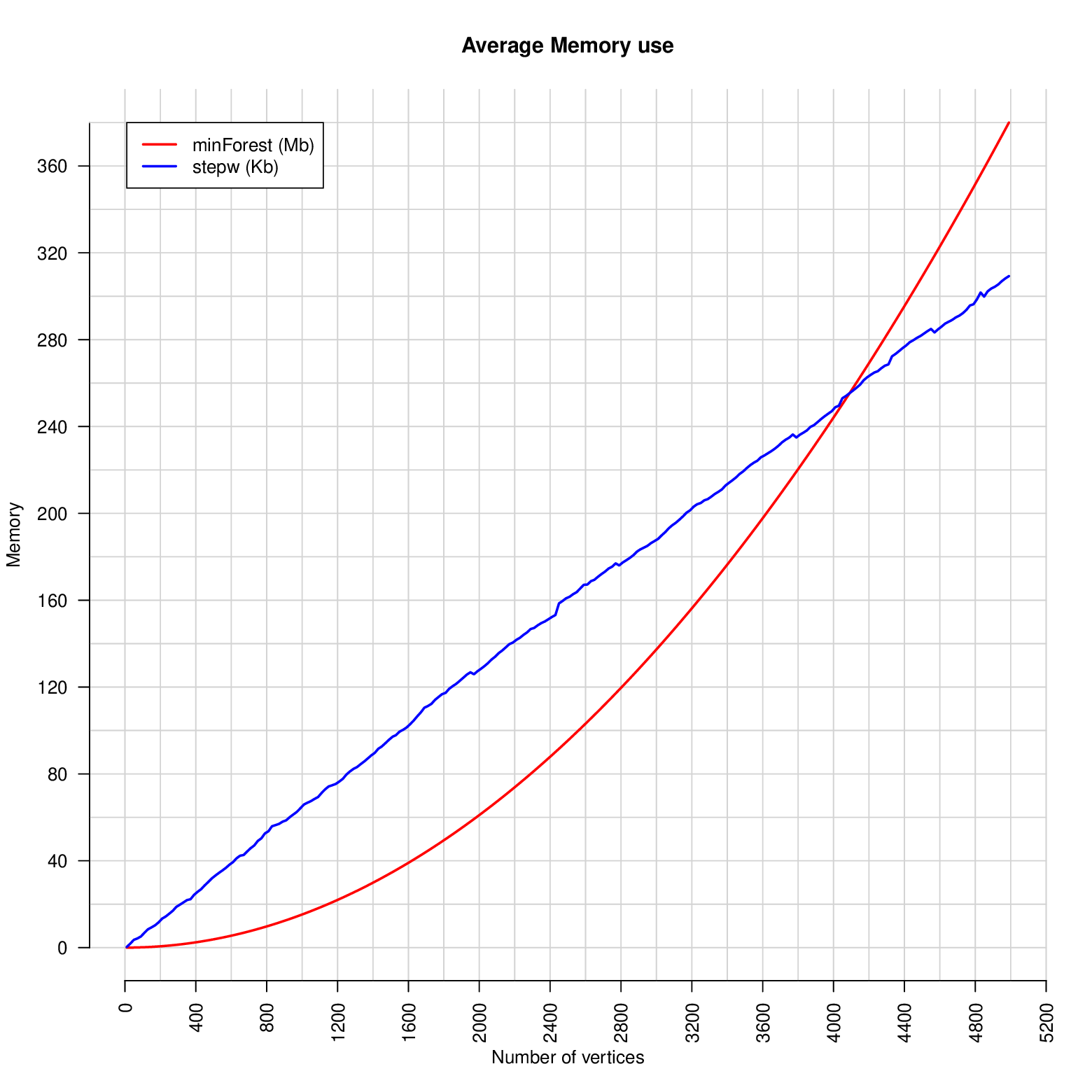}
\includegraphics[width=.48\linewidth]{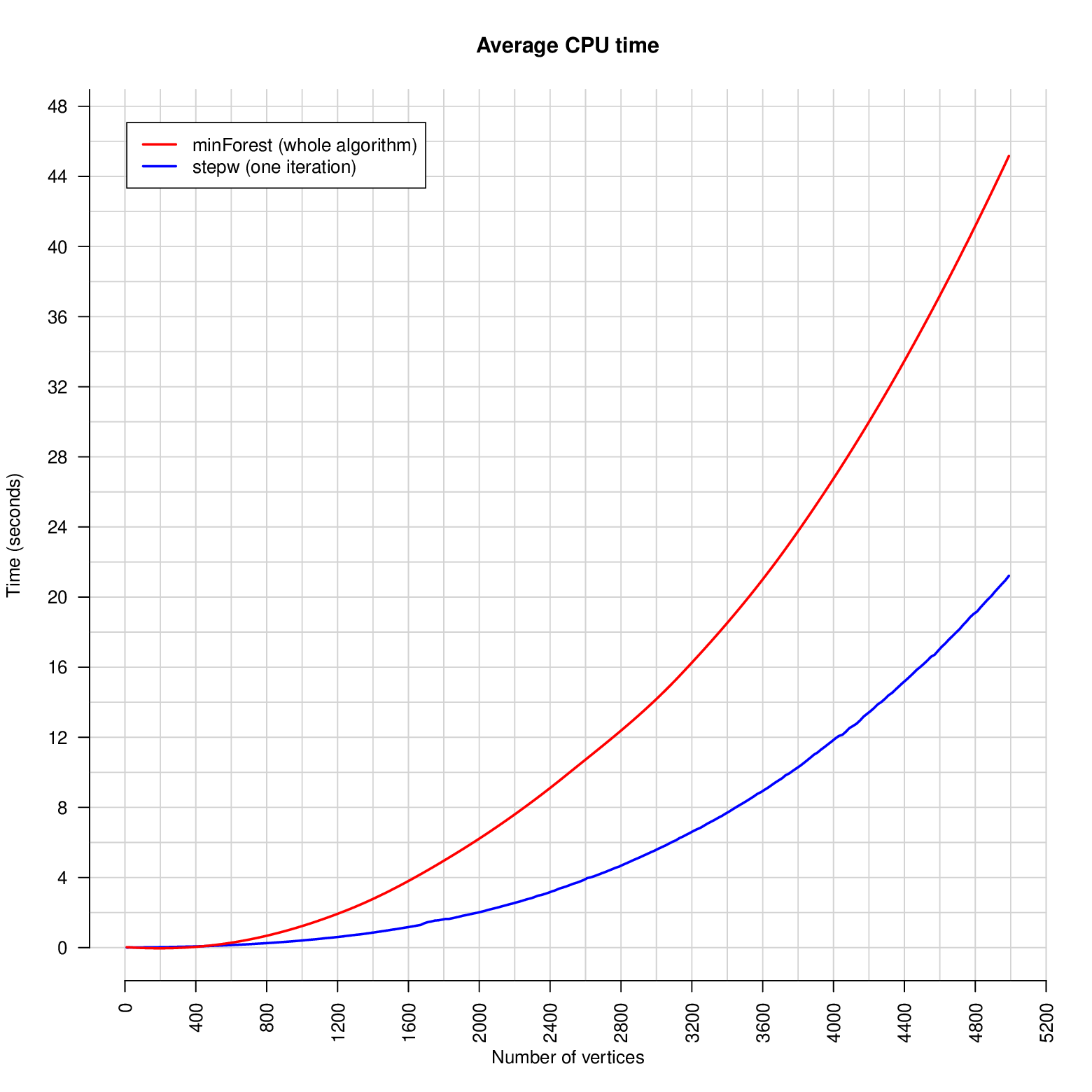}
\caption{{\bf Computational performance}. Memory use and CPU time. The curve for the CPU time of the {\tt stepw} function reflects an average iteration, while for the {\tt minForest} function it is for the whole algorithm.}
\label{fig:perf}
\end{figure}

Additionally, we performed a simulation study where the functions \verb@minForest@ and \verb@stepw@ were evaluated. Data sets from multivariate normal distributions were generated with 10 to 5,000 variables (at intervals of 20). Each of these data sets were used
to infer the minimum spanning forest and a more complex decomposable graph starting from this forest. The computational performance is showed in Figure~\ref{fig:perf}. As the \verb@stepw@ function depends on the final complexity of the model (number of edges added to the model), the CPU time shown is for one (average) iteration only. The memory demand for the \verb@minForest@ function grows quadratically in the number of vertices, while the memory demand for the \verb@stepw@ function grows linearly in the number of vertices. The CPU time also grows quadratically in both functions.

In conclusion, the \verb@minForest@ function runs much faster than the \verb@stepw@, allowing the selection of a skeleton of the model in a forest-like structure (sparse) in a very short time. We remark that it appears to be much more efficient to start the \verb@stepw@ function from the forest found by \verb@minForest@, rather than from the null model.

\section{Analysis of the graphical structure}
\label{sec:analysis}
When high-dimensional models are studied, plotting the model found is probably not the best way to visualise the result, as can be seen in Figure~\ref{fig:final}. The plotting algorithm used here is time consuming, and may need a large number of iterations to generate a graph with no edges crossing each other. Therefore different ways to analyse the resulting model have to be used. The package contains a number of functions which could be useful for this task:
\begin{itemize}
    \item \verb@adjMat@: Returns the adjacency matrix based on a list of edges.
    \item \verb@fit@: Calculate a model's -2$\times$log-likelihood, AIC, and BIC.
    \item \verb@Degree@: Returns the degree of a set of vertices.
    \item \verb@DFS@: Finds all vertices reachable from one specific vertex (assuming that there are no cycles).
    \item \verb@findEd@: Finds all add-eligible edges to a given triangulated graph, preserving triangularity. In the case of a mixed graph, forbidden edges are not returned.
    \item \verb@neighbours@: Returns all vertices with a direct connection with a vertex \verb@v@.
    \item \verb@MCS@: Returns a perfect numbering of the edges.
    \item \verb@modelDim@: Calculates the number of free parameters corresponding to each edge.
    \item \verb@modelFormula@: Finds the model's formula \cite[pages 213-216]{lauritzen1996}.
    \item \verb@neighbourhood@: Finds the set of vertices with up to a given distance from a given vertex.
    \item \verb@perfSets@: Finds a perfect sequence, returning the cliques, histories, residuals, and separators of a given triangulated graph.
    \item \verb@shortPath@: Calculates the shortest path between a vertex \verb@v@ and all other vertices.
    \item \verb@SubGraph@: Based on a list of vertices, generates a subgraph.
    \item \verb@summary@: Gives details about the model's structure.
    \item \verb@jTree@: Finds a junction tree of a graph.
    \item \verb@CI.test@: Calculates the deviance and adjusted degrees of freedom for the conditional independence test.
\end{itemize}

\begin{figure}[!ht]
\centering
\includegraphics[height=5.1cm,width=5.1cm]{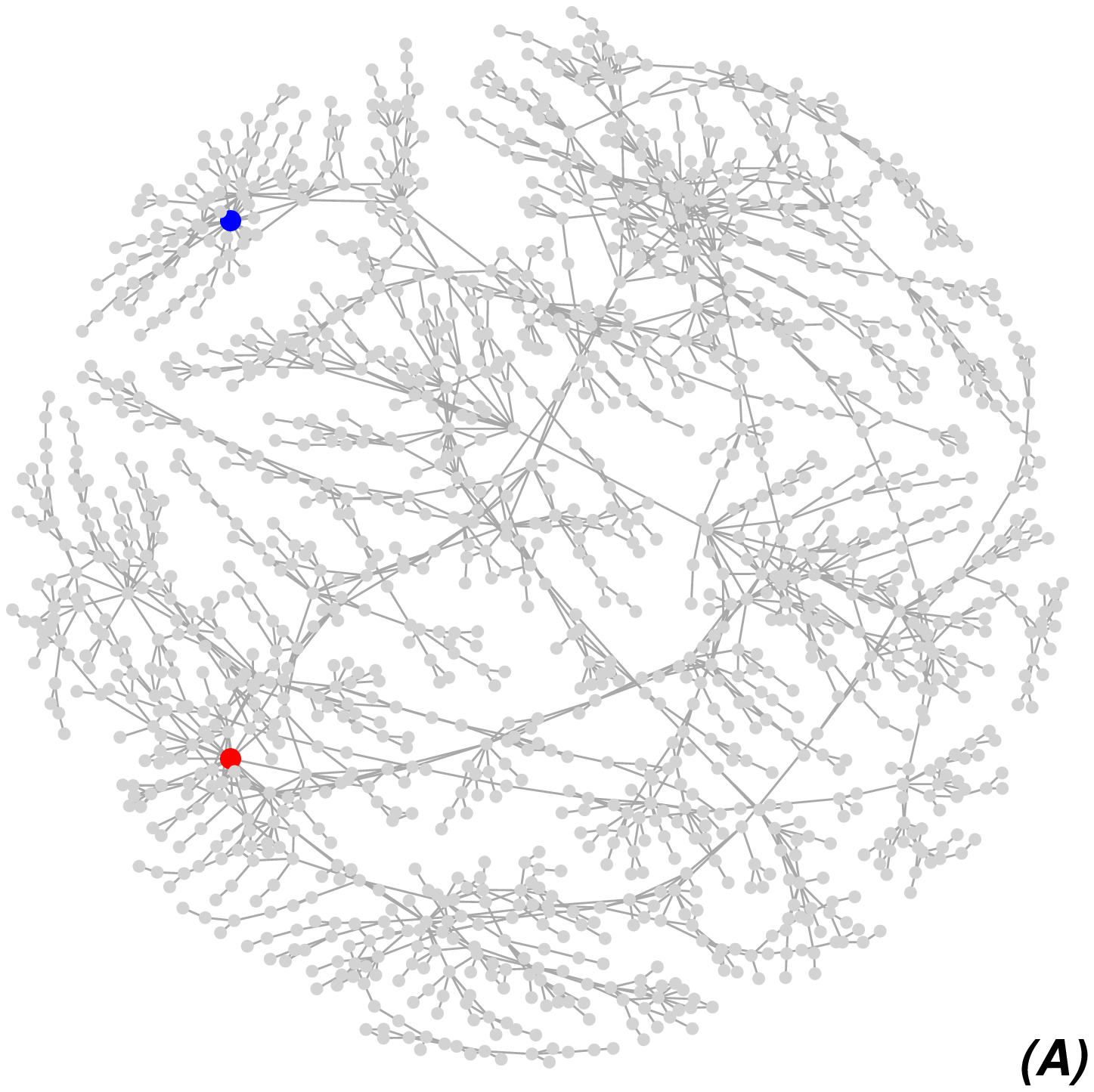}
\includegraphics[height=5.1cm,width=5.1cm]{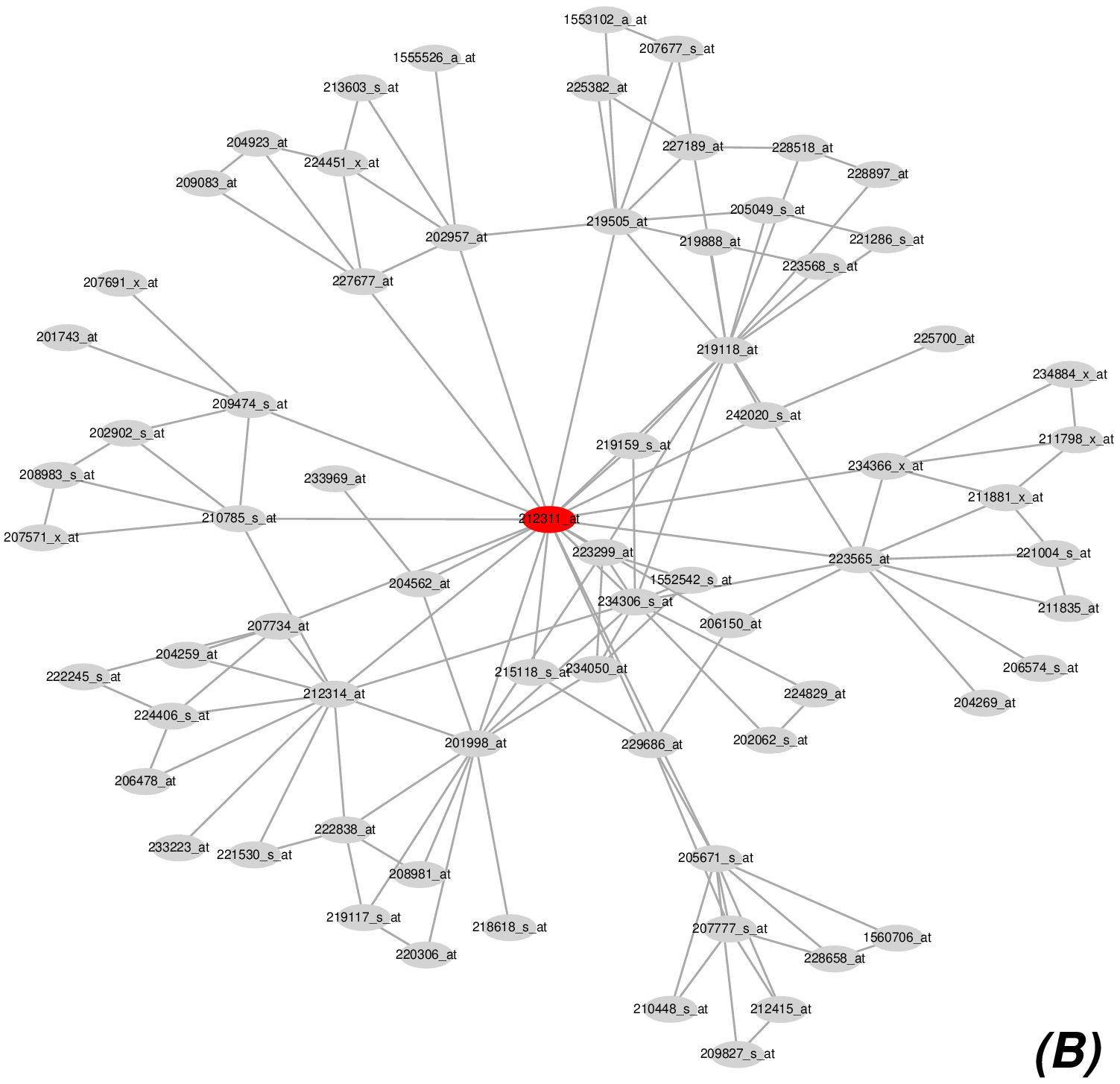}
\includegraphics[height=5.1cm,width=5.1cm]{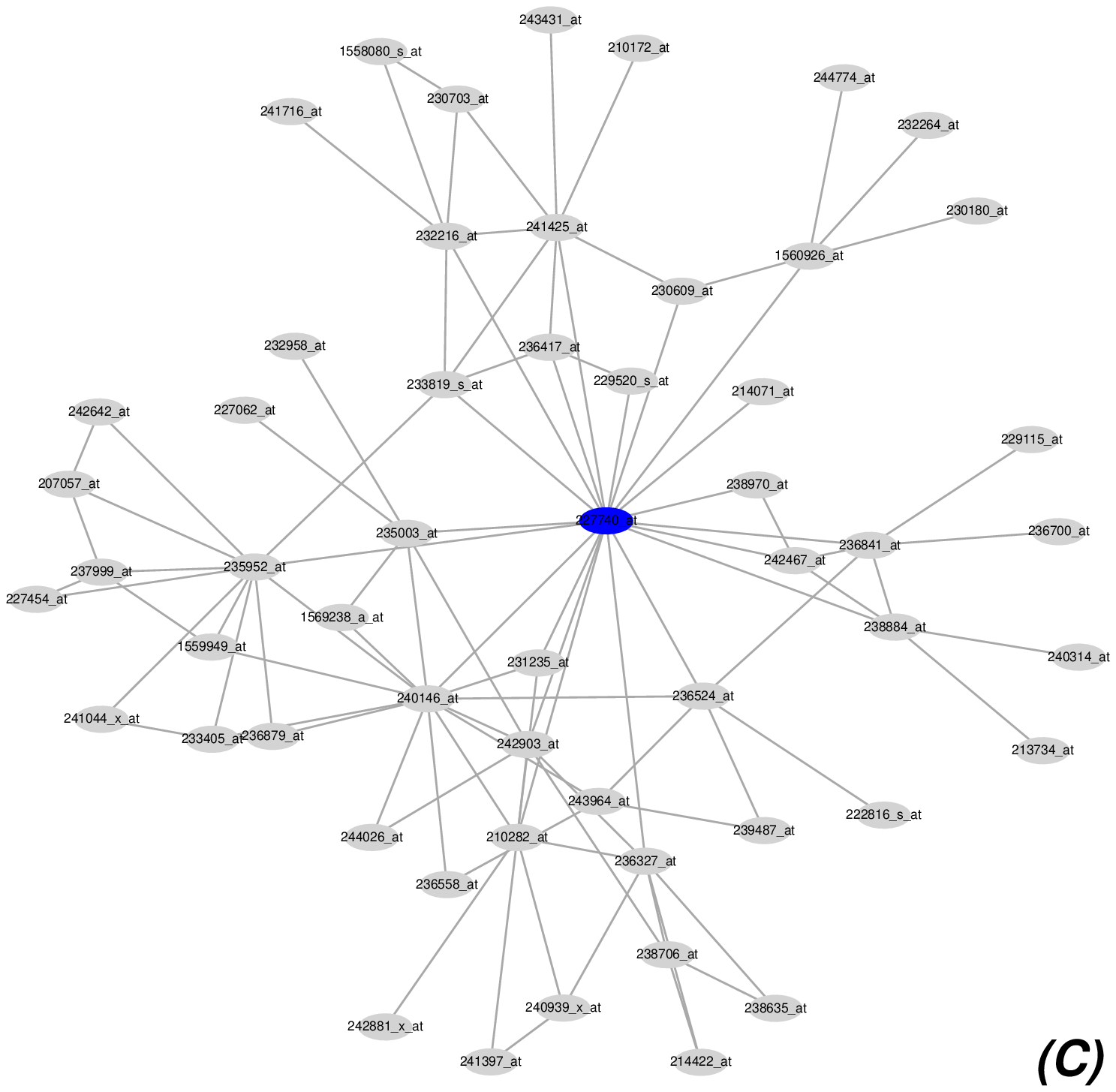}
\caption{{\bf Zooming in specific regions of a large graph}. (A) The final graph of the periodontitis data, highlighting vertices 670 (red) and 1,123 (blue). (B) The neighbourhood of vertex 670, including only vertices within a radius of two. (C) The neighbourhood of vertex 1,123, including only vertices within a radius of two.}
\label{fig:neigh}
\end{figure}

The degree of the periodontitis model shows two vertices with high number of direct neighbours (degree 20). These vertices could be ``hubs", with important action in the network. We can also see that there is a high number of ``leafs" in the graph, as 382 (24.72\%) vertices present only one edge.
\begin{verbatim}
R> table(Degree(periodontitisForward))
  1   2   3   4   5  6  7  8  9 10 11 12 13 14 15 16 17 20
382 379 249 185 116 83 53 28 18 17 13  6  5  6  1  1  1  2
\end{verbatim}
We can zoom in the neighbourhood (up to the second neighbour, for example) of these two ``hubs".
\begin{verbatim}
R> vertices <- which(Degree(periodontitisForward)==20)
R> neigh670 <- neighbourhood(periodontitisForward,orig=vertices[1],rad=2)
R> pos <- plot(periodontitisForward,numIter=4000,vert.labels=FALSE,
+              vert.radii=.006,vert.hl=vertices,col.hl=c("red","blue"))
R> plot(periodontitisForward,vert=neigh670$v[,1],numIter=1000,
+       vert.hl=vertices[1],vert.radii=.013,col.hl="red",cex.vert.label=.4)
R> plot(periodontitisForward,vert=neigh1123$v[,1],numIter=1000,
+       vert.hl=vertices[2],vert.radii=.013,col.hl="blue",cex.vert.label=.4)
\end{verbatim}
From Figure~\ref{fig:neigh} we see that restricting the plot to a smaller neighbourhood
allows details to become more visible. The same plot could also be produced using the functions \verb@neighbours@ and \verb@SubGraph@.

For the HapMap data, the degree shows that we actually have a high number of isolated vertices (104, 31.14\%). The general structure of the model is very close to a forest, as only 5 edges are added by the \verb@stepw@ function. The vertices with highest degree have 5 edges.
\begin{verbatim}
R> table(Degree(HapMapForward))
  0   1   2   3   4   5
104 133  58  22  12   5
\end{verbatim}

The log-likelihood, BIC, and AIC of a model can be obtained using the function \verb@fit@. For example, if we are interested in test in a mixed model if the structure of covariance is homogeneous or heterogeneous, we could use the this information.
\begin{verbatim}
R> fit(edges=irisForward,dataset=iris,homog=FALSE)
Number of parameters   -2*Log-likelihood        AIC        BIC
             39.0000            410.1435   488.1435   605.5582
R> fit(edges=irisForward,dataset=iris,homog=TRUE)
Number of parameters   -2*Log-likelihood        AIC        BIC
             22.0000            551.3188   595.3188   661.5528
\end{verbatim}

The \verb@perfSets@ function finds the cliques structure of the graph, also returning the lists of separators, histories, and residuals, as described in
\cite{lauritzen1996} (pages 14-15). If the graph is not (strongly) decomposable, the function returns the value zero. For the periodontitis data we have
\begin{verbatim}
R> sets <- perfSets(periodontitisForward)
# only the first two elements of each are shown here
List of 4
    $ cliques :List of 1395
     ..$ : int [1:2] 1 335
     ..$ : int [1:3] 394 335 474
    $ histories :List of 1395
     ..$ : int [1:2] 1 335
     ..$ : int [1:4] 1 335 394 474
    $ separators :List of 1395
     ..$ : NULL
     ..$ : int 335
    $ residuals :List of 1395
     ..$ : int [1:2] 1 335
     ..$ : int [1:2] 394 474
\end{verbatim}

The \verb@shortPath@ function returns the shortest path length between vertices in the graph (considering that each edge has length one). If two vertices are not connected, i.e., there is no path between them, it is returned \verb@Inf@. In the periodontitis data, the vertex 670 has a direct connection with 20 other vertices, and the most far vertex has a distance of 26 from it, while the graph's diameter (longest shortest path) is 46. Note that vertex 670 has a distance of 0 to itself.
\begin{verbatim}
R> table(shortPath(periodontitisForward,v=670))
 0   1   2   3   4   5   6   7   8   9  10  11  12  13  14  15  16  17
 1  20  50  58  72  78  84  77  80  97 134 130 126 116  98  91  80  70
18  19  20  21  22  23  24  25  26
48  15   4   4   4   3   3   1   1
\end{verbatim}

\section{Plotting a graphical model}
\label{sec:plot}
Some examples of plots generated by the \verb@plot.gRapHD@ function have been given above. The function uses S3 method for the class \verb@gRapHD@, so the regular \verb@plot@ function can be used. As default, when a \verb@gRapHD@ object is plotted, all discrete variables are pictured as black circles, and the continuous variables as grey circles. But the function is flexible, and the user can define different colours, shapes, and sizes for each vertex. The \cite{fruchterman1991} algorithm is used to place the vertices in the plotting area \cite{csardi2006}. The algorithm is iterative and uses attractive and repulsive forces for placing the vertices. This technique is
time consuming and cannot guarantee a clear plot, as the one in Figure~\ref{fig:final} (A), which used 4,000 iterations, a number not sufficient to untangle it.

\begin{figure}[!ht]
\centering
\includegraphics[width=.55\linewidth]{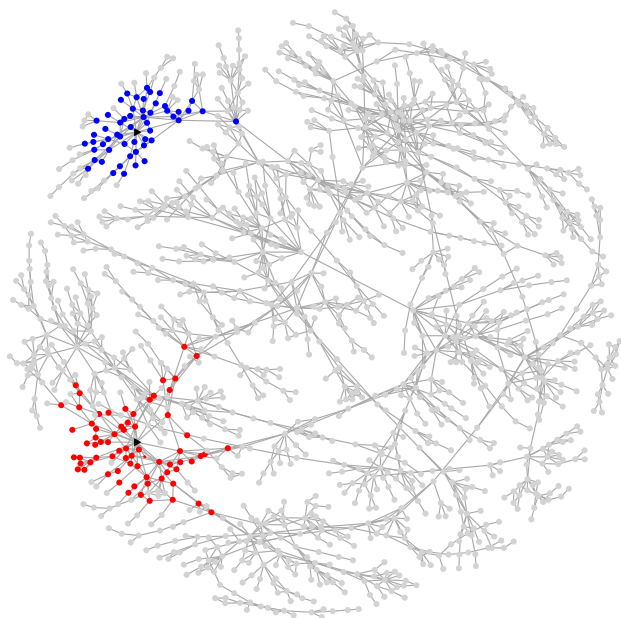}
\caption{{\bf Highlighting specific regions in a complex graph}. The vertices 670 and 1,123 are pictured in black as triangles, while the respective neighbourhoods are picture in red and blue.}
\label{fig:whole_neigh}
\end{figure}

The more complex a graph is, the more difficult is its visualisation. For this reason, the \verb@plot.gRapHD@ function has a number of optional parameters that allow the user to manipulate the appearance of the plot. It is possible to plot only the edges, or not to label the vertices; to highlight some vertices with different colours, shapes, and/or sizes. For example we could show in the graph where the neighbourhood of the vertices 670 and 1,123 are, for the periodontitis data, as shown in Figure~\ref{fig:whole_neigh}.

\begin{verbatim}
R> hlv <- c(670,1123,neigh670$v[-1,1],neigh1123$v[-1,1])
R> hlc <- rep(c("black","red","blue"),c(2,length(neigh670$v[-1,1]),
+             length(neigh1123$v[-1,1])))
R> vs <- rep(0.005,periodontitisForward$p)
R> vs[c(neigh670$v[,1],neigh1123$v[,1])] <- .008
R> vs[c(670,1123)] <- .01
R> sb <- rep(1,periodontitisForward$p)
R> sb[c(670,1123)] <- 3
R> vs[901]<-0
R> plot(periodontitisForward,coord=pos,numIter=0,vert.hl=hlv,col.hl=hlc,
+       vert.labels=FALSE,vert.radii=vs,symbol.vert=sb)
\end{verbatim}

\begin{figure}[!ht]
\centering
\includegraphics[width=.55\linewidth]{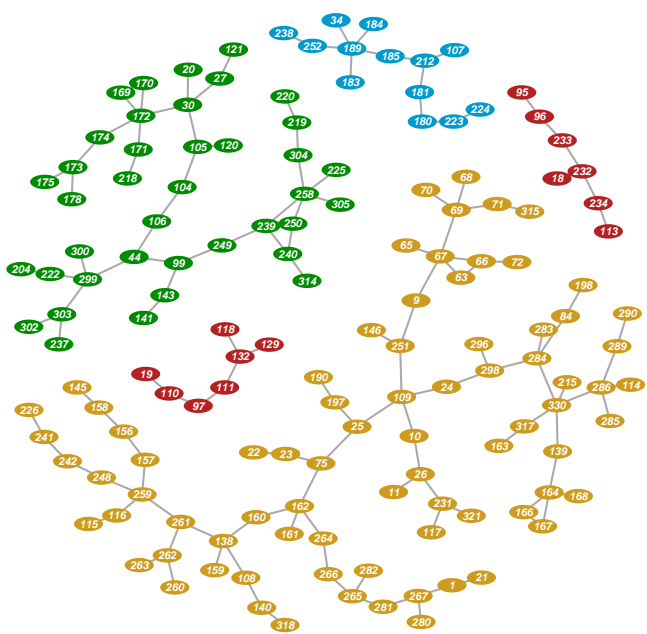}
\caption{{\bf Largest components in the HapMap graph}. The components with more than 5 vertices are plotted in different colours - red (7 vertices); blue (13 vertices); green (39 vertices); and yellow (79 vertices).}
\label{fig:comp_HapMap}
\end{figure}

The code below can be used to identify isolated components in a graph. The largest components in the HapMap data are displayed in Figure~\ref{fig:comp_HapMap}.

\begin{verbatim}
R> sp <- shortPath(HapMapForward)
R> comp <- rep(0,HapMapForward$p)
R> i <- 0
R> while(length(which(comp==0))>0) {
R>   i <- i + 1
R>   if (comp[i] == 0) {
R>     ind <- which(sp[i,]<HapMapForward$p) #finite
R>     comp[ind] <- max(comp) + 1
R>   }
R> }
R> v <- c(which(comp==18),which(comp==21),which(comp==15),
          which(comp==1),which(comp==9))
R> col <- rep(colours()[c(133,124,258,150)],c(14,13,39,79))
R> plot(HapMapForward,vert=v,lwd.ed=3,vert.hl=v,symbol.vert=rep(0,length(v)),
+   vert.radii=rep(.012,length(v)),col.hl=col,numIter=3000,cex.vert.label=.7)
\end{verbatim}

\section{Concluding remarks}
\label{sec:disc}
We have described an \proglang{R} package for efficient selection of high-dimensional undirected graphical models, with functions not available in other packages. The objective of the package is to provide an efficient way to search for models in the classes of forests and decomposable graphs (discrete, continuous, and mixed). The computational performance depends on the model complexity, where sparse models demand much less resources. It is easy to convert \verb@gRapHD@ objects into graph or model objects supported by other packages; it is also easy to extract information directly from them. There is a technical limitation of 65,000 variables, due to numerical representation.

\section*{Acknowledgements}
Gabriel C. G. Abreu is financed by SABRETRAIN Project, funded by the Marie Curie Host Fellowships for Early Stage Research Training, as part of the $6^{th}$ Framework Programme of the European Commission. R.L. was partially supported by the project ``Metabolic programming in Foetal Life'', Danish Research Agency, Ministry of Science Technology and Innovation.


\begin{thebibliography}{10}

\bibitem{bollobas2000}
Bela Bollobas.
\newblock {\em Modern Graph Theory}.
\newblock Springer, 2000.

\bibitem{bondy2008}
J.~A. Bondy and U.~S.~R. Murty.
\newblock {\em Graph Theory}.
\newblock Springer, 2008.

\bibitem{chow1968}
C.~K. Chow and C.~N. Liu.
\newblock Approximating discrete probability distributions with dependence
  trees.
\newblock {\em IEEE Transactions on Information Theory}, 14(3):462--467, May
  1968.

\bibitem{csardi2006}
Gabor Csardi and Tamas Nepusz.
\newblock The \pkg{igraph} software package for complex network research.
\newblock {\em InterJournal}, Complex Systems:1695, 2006.

\bibitem{darroch1980}
J.~N. Darroch, S.~L. Lauritzen, and T.~P. Speed.
\newblock Markov fields and log-linear interaction models for contingency
  tables.
\newblock {\em Annals of Statistics}, 8:522--539, 1980.

\bibitem{demmer2008}
Ryan~T. Demmer, Jan~H. Behle, Dana~L. Wolf, Martin Handfield, Moritz Kebschull,
  Romanita Celenti, Paul Pavlidis, and Panos~N. Papapanou.
\newblock Transcriptomes in healthy and diseased gingival tissues.
\newblock {\em Journal of Periodontology}, 79(79):2112--2124, 2008.

\bibitem{dempster1972}
A.P. Dempster.
\newblock Covariance selection.
\newblock {\em Biometrics}, 29:157--175, 1972.

\bibitem{dhamodaran2008}
S.~Dhamodaran, A.~Saad, and D.~Fink.
\newblock Application of network theory for the description of nanocluster
  distributions in ion track electronics.
\newblock {\em Radiation Effects and Defects in Solids}, 163(9):749--759,
  September 2008.

\bibitem{dorogovtsev2003}
S.N. Dorogovtsev and J.F.F. Mendes.
\newblock {\em Evolution of Networks: From Biological Networks to the Internet
  and WWW}.
\newblock Oxford University Press, 2003.

\bibitem{dunne2002}
Jennifer~A. Dunne, Richard~J. Williams, and Neo~D. Martinez.
\newblock Food-web structure and network theory: The role of connectance and
  size.
\newblock {\em Proceedings of the National Academy of Sciences},
  99(20):12917--12922, 2002.

\bibitem{edwards2000}
David Edwards.
\newblock {\em Introduction to Graphical Modelling}.
\newblock Springer-Verlag New York Inc., 2000.

\bibitem{edwards2010}
David Edwards, Gabriel Coelho~Gon\c{c}alves Abreu, and Rodrigo Labouriau.
\newblock Selecting high-dimensional mixed graphical models using minimal aic
  or bic forests.
\newblock {\em BMC Bioinformatics}, 11(1):18, 2010.

\bibitem{faith2007}
Jeremiah~J Faith, Boris Hayete, Joshua~T Thaden, Ilaria Mogno, Jamey
  Wierzbowski, Guillaume Cottarel, Simon Kasif, James~J Collins, and Timothy~S
  Gardner.
\newblock Large-scale mapping and validation of escherichia coli
  transcriptional regulation from a compendium of expression profiles.
\newblock {\em PLoS Biology}, 5(1):e8, 01 2007.

\bibitem{fruchterman1991}
Thomas M.~J. Fruchterman and Edward~M. Reingold.
\newblock Graph drawing by force-directed placement.
\newblock {\em Software: Practice and Experience}, 21(11):1129--1164, 1991.

\bibitem{gautier2004}
Laurent Gautier, Leslie Cope, Benjamin~M. Bolstad, and Rafael~A. Irizarry.
\newblock \pkg{affy}---analysis of affymetrix genechip data at the probe level.
\newblock {\em Bioinformatics}, 20(3):307--315, 2004.

\bibitem{goodman1973}
Leo~A. Goodman.
\newblock The analysis of multidimensional contingency tables when some
  variables are posterior to others: A modified path analysis approach.
\newblock {\em Biometrika}, 60(1):179--192, April 1973.

\bibitem{krause2007}
J.~Krause, D.~P. Croft, and R.~James.
\newblock Social network theory in the behavioural sciences: Potential
  applications.
\newblock {\em Behavioral Ecology and Sociobiology}, 62(1):15--27, November
  2007.

\bibitem{lauritzen1989}
S.~L. Lauritzen and N.~Wermuth.
\newblock Graphical models for associations between variables, some of which
  are qualitative and some quantitative.
\newblock {\em Annals of Statistics}, 17(1):31--57, 1989.

\bibitem{lauritzen1996}
Steffen~L. Lauritzen.
\newblock {\em Graphical Models}.
\newblock Oxford University Press, 1996.

\bibitem{HapMap2003}
{The International HapMap Consortium}.
\newblock The international hapmap project.
\newblock {\em Nature}, 426(6968):789--796, December 2003.

\bibitem{whittaker1990}
Joe Whittaker.
\newblock {\em Graphical Models in Applied Multivariate Statistics}.
\newblock John Wiley and Sons, 1990.

\bibitem{yosef2009}
Nir Yosef, Martin Kupiec, Eytan Ruppin, and Roded Sharan.
\newblock A complex-centric view of protein network evolution.
\newblock {\em Nucleic Acids Research}, 37(12):e88, July 2009.

\end{thebibliography}

\end{document}